\documentclass[manuscript,screen]{acmart}

\AtBeginDocument{%
  \providecommand\BibTeX{{%
    \normalfont B\kern-0.5em{\scshape i\kern-0.25em b}\kern-0.8em\TeX}}}

\setcopyright{acmcopyright}
\copyrightyear{2018}
\acmYear{2018}
\acmDOI{10.1145/1122445.1122456}

\acmConference[Woodstock '18]{Woodstock '18: ACM Symposium on Neural
  Gaze Detection}{June 03--05, 2018}{Woodstock, NY}
\acmBooktitle{Woodstock '18: ACM Symposium on Neural Gaze Detection,
  June 03--05, 2018, Woodstock, NY}
\acmPrice{15.00}
\acmISBN{978-1-4503-XXXX-X/18/06}

\usepackage{xspace}
\usepackage{caption}
\usepackage{subcaption}

\begin{document}

\title{The Road Less Travelled: Trying And Failing To Generate Walking Simulators}

\author{Michael Cook}
\email{mike@possibilityspace.org}
\affiliation{%
  \institution{Queen Mary University of London}
  \city{London}
  \country{UK}
}


\renewcommand{\shortauthors}{Cook}

\newcommand{\angelina}{\textit{ANGELINA}\xspace}

\begin{abstract}
Automated game design is a rapidly growing area of research, yet many aspects of game design lie largely unexamined still, as most systems focus on two-dimensional games with clear objectives and goal-oriented gameplay. This paper describes several attempts to build an automated game designer for 3D games more focused on space, atmosphere and experience. We describe our attempts to build these systems, why they failed, and what steps and future work we believe would be useful for future attempts by others.
\end{abstract}



\maketitle

\section{Introduction}
Automated game design (AGD) is the theory and practice of building systems that take on responsibilities in any part of the game design process. It overlaps with computational creativity as well as procedural content generation, and has roots stretching back long before digital games research had begun in the form we know it today \cite{metagame}.

In \cite{smithcook} Cook and Smith offer a critique of the field, suggesting that the history of AGD research, at the time of writing in 2015, was primarily focused on the generation of rules for games, and limited to goal-oriented games with clear objective functions for winning. They write:

\begin{quote}
\textit{This mechanics-first view on games is unnecessarily limiting, stifling the creative potential for AGD and restricting the kinds of games that can be automatically designed to ones that have well-defined, simple rule systems.}
\end{quote}

More than half a decade on from the publication of this work, and most of its points still hold true of AGD research today. This is not in itself a flaw in the research being done -- it is still valuable, and the field is progressing and creating many new and exciting systems \cite{gemini,sarkar}. Yet there remains a need to expand beyond this, to create the ``new kinds of play experience'' that Cook and Smith talk about, to expand the horizons of AGD as a research field, and most importantly to expand the scope of how AI interacts with, improves and changes games as a creative medium.

One possible reason that AGD has not expanded into the areas suggested by Smith and Cook is that it is a much harder space to do AGD research in. We claim this from personal experience, as we ourselves attempted to build systems that designed `puzzle-boxes' -- games about exploring spaces, appreciating their architecture and design, and finding delightful or surprising interactions and artefacts within. Each of these systems failed to yield publishable work or to progress beyond early prototypes, and in some cases failed even to yield playable and interesting results. 

Despite this, we have learned many things both from trying to build these systems, and thinking about the challenges and failures met along the way. These are not impossible problems to solve, and failure is not inevitable. We know other researchers are considering these problems now, and look forward to their results and insights. In the spirit of the Reflections track, this paper attempts to look back on our own attempts, identify why they failed and what insight we gained through this process, and point to new approaches and opportunities in AGD as a result, that hopefully others will find more success in tackling.

\section{Background}
\subsection{Variety in AGD Research}
Although the field of automated game design (AGD) is relatively small, there are an increasing number of researchers working in this area, developing a wide range of AGD systems, covering genres such as arcade games \cite{angelina1}, boardgames \cite{ludi}, political newsgames \cite{gom} and platformers \cite{sarkar}.

Despite this valuable existing work, much of the thematic and mechanical space represented by both physical and digital games lies unexplored. There are, to our knowledge, no AI systems that attempt to design rhythm games, sports games, dating games or open world RPGs, for example. In \cite{smithcook} the authors suggest that AGD research has largely restricted itself to design spaces where there are fairly precise measures for success in gameplay -- such as score, in the case of arcade games, or reaching the end of a level. The domains are also largely defined by their rules: that is, if the rules describe an interesting set of systems, then the game is considered successful. This might justify putting less emphasis on other aspects of the design, such as art or music.

This is not to say that these factors have been ignored entirely, nor that AGD researchers are still uninterested or ignorant of their existence. Several systems do engage with other aspects -- systems going as far back as \cite{nelsonmateas} have incorporate visual content, for example. Many AGD researchers are also deeply interested in expanding their research to cover these new areas.

We believe that several factors influence the direction of AGD research to date, in addition to those that Cook and Smith identify in their 2015 paper. 

\paragraph{Scope Size} Many of the domains commonly chosen can be developed as small, self-contained games that take place in small spaces, usually a single level and sometimes a single screen. This makes prototyping AGD systems easier, playtesting games easier, and also makes the development of the underlying game engine easier too. By contrast, developing an AGD system that designs 50-hour RPGs would be complex to build, produce games that took a long time to playtest, and would also involve a large game development task if a suitable existing game engine cannot be repurposed.

\paragraph{Known Blockers} In many cases unexplored genres represent known, complex AI challenges even prior to any AGD exploration of the area. For example, the design of visual novels necessitates solving problems such as narrative design or dialogue generation. These tasks are already known to be frontier problems in AI research, and although investigating the genre might reveal new and exciting related problems, it is likely that new researchers seeking to tackle these challenges might instead be working on research in, for example, narrative generation. 

\paragraph{Lack of Experience} Researchers bring their own experience and history to their work, which influences both their interests and the areas they feel confident and knowledgeable about. For example, the proliferation of Zelda and Mario benchmarks in Game AI research is influenced by the age and regional background of researchers within the community. Similarly, AGD researchers who have little experience in a particular genre (such as, for example, racing sims) might therefore feel less confident building AI systems which design in those genres. AGD is a research discipline that requires researchers also, to some extent, think about and practice game design themselves, and thus personal experience and preference naturally factor in.

\subsection{Walking Simulators}
Genre is a complicated concept, and common categories for games are often unfit for purpose or rooted in older ideas about games from the 1980s. Genres offered on Steam, one of the largest online stores for purchasing games, are either unrelated to the game's content, such as \textit{Free to Play}, hopelessly vague, such as \textit{Action}, or constantly in conflict over their definition, such as \textit{Roguelike}. 

In \cite{smithcook} the authors isolate a more recently-defined genre of games called \textit{walking simulators}. This term has been used both positively and negatively and was a source of conflict at the time, and so Cook and Smith use an alternative term, \textit{secret box game}, instead. In this paper we use the term walking simulator, as it has now been largely adopted as the default (and positive) term to describe the genre, and is a little more immediately descriptive.

Walking simulators are not strictly defined, but usually have a few of the following characteristics: they are played from a first-person perspective, with the primary method of control being walking and usually containing few controls besides this. The player's goals are often to simply explore a space and see what it contains, but there may also be narrative to uncover, or puzzles to solve. They typically do not contain any reflex challenges. Traditional examples of walking simulators include \textit{Dear Esther}, \textit{Firewatch} and \textit{Tacoma}. However, games such as \textit{A Short Hike}, \textit{The Occupation} or \textit{Umurangi Generation} might also be considered to have one foot in the genre, even though they contain conflicting elements. The \textit{Bitsy} game engine also produces games similar in tone to walking simulators, but viewed from a top-down 2D perspective.

Walking simulators are identified as a target for AGD research by Cook and Smith because many of their most definitive exemplars, such as \textit{Dear Esther}, are almost devoid of rules or goals, and emphasise aesthetics, design and atmosphere. Considering the factors that make AGD difficult that we identified in the previous section, walking simulators fare well. In terms of \textit{scope size}, walking simulators scale well to various lengths, from several hours down to a few minutes. In terms of \textit{Lack of Experience}, walking simulators are a relatively popular modern genre that includes many well-known blockbuster games that many researchers will be aware of. More importantly, though, they build on game skills common to other genres - such as first-person shooters, or action games. Even in the \textit{Known Blocker} category, walking simulators can be viewed favourably. Although the genre is complex and contains many challenges (as we will discuss shortly) these challenges are not especially well-understood or being actively studied elsewhere.

\subsection{Our Research}
In order to provide some further context for the discussion about projects contained in the remainder of this paper, we will briefly outline our normal mode of working, the tools used, and the approach we take to AGD research, to help provide insight into how we decided to undertake these experiments and, perhaps, why they ended up not succeeding. This also acts as a sort of overview of the following sections.

Our original motivation for working in this area was the paper by Cook and Smith. Typically, we begin work in a new area of AGD research by surveying the area by looking at games from the genre, and making small prototypes by hand. This involved looking at popular walking simulators like \textit{Dear Esther}, but also at smaller examples, in particular at KO-OP Mode's \textit{Weird Forest} Unity tutorial that involves a minimalist forest scene as a sample. This scene is very small, but has many of the key properties of a walking simulator -- a space to explore, hidden things to find, and strong overall art direction. This made it a good `inspiring example' for our own work.

\begin{figure}
\includegraphics[width=\columnwidth]{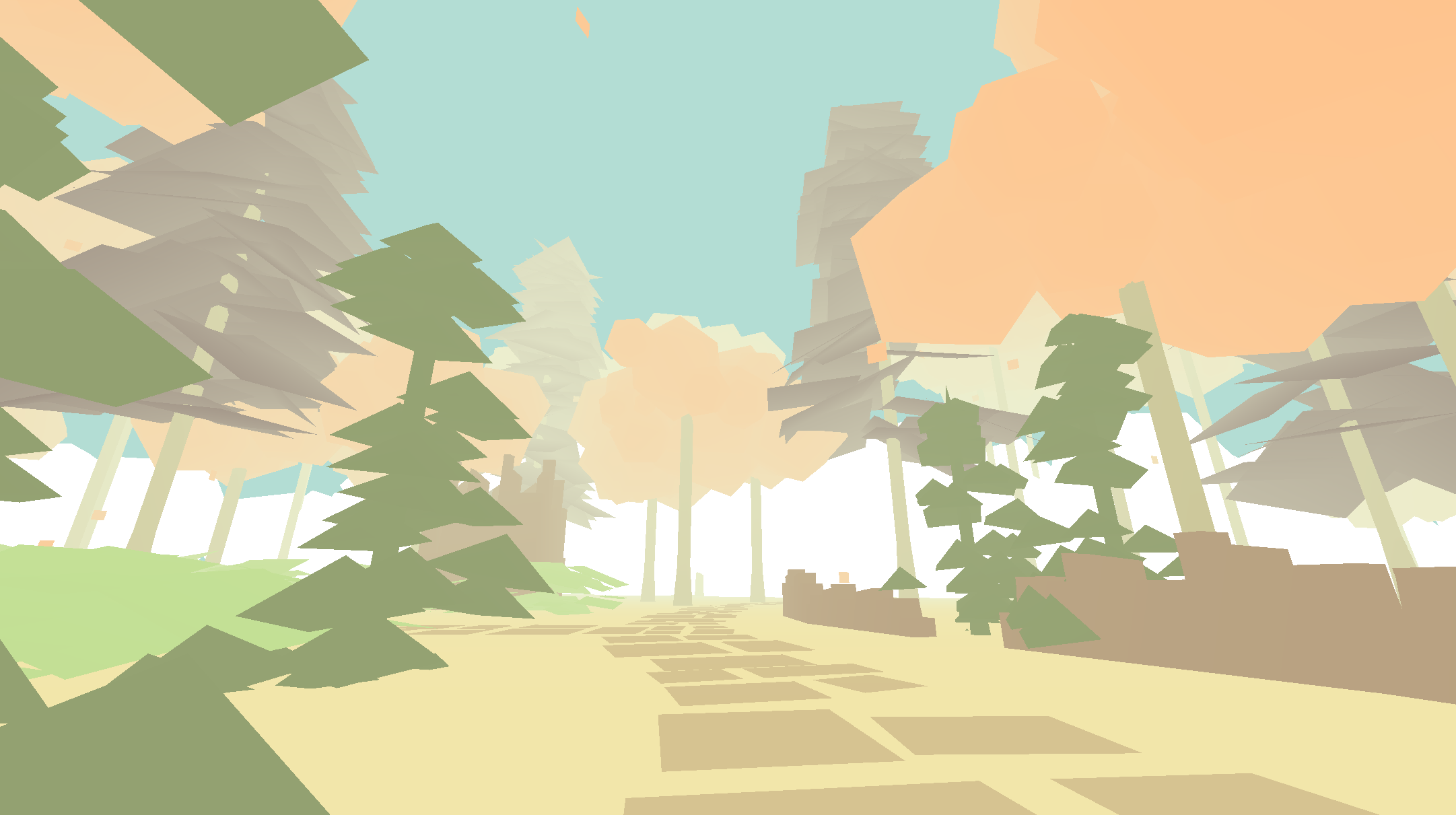}
\caption{A screenshot of KO-OP Mode's `Weird Forest' tutorial.}
\end{figure}

We undertook three different projects to probe the task of using AGD techniques to generate walking simulators. First, we began by building a system inspired by our most recent AGD system at the time, \angelina. This was an exploratory project to examine what the difficulties in building an automated walking simulator designed would be like, with the \textit{Weird Forest} as an inspiring target. The second project was a technical investigation into alternative forms of evaluation, using vision as the primary driver of evaluation. The third project was an attempt to refocus automatic walking simulator design on a simplified target domain, with more supporting design.

All of our research was conducted in Unity. At the time, we were learning how to use Unity in order to broaden our skillset, but even our decision to use Unity was influenced by a desire to use and work with tools that are used by everyday game developers. More and less complex tools exist, but Unity felt like a good target environment that represented real design tools, and therefore the potential to build systems that might be useful to real designers. Unity also simplified distribution of our results, and many of our inspiring examples were also designed in Unity, which supported this decision further.

In the following sections we will discuss each of these three projects, as well as some of the difficulties we encountered in their development. At the end, we'll discuss our reflections on the whole effort, as well as explaining our decision to put these efforts on hold.

\section{System 1 - Full Automation Of Explorable 3D Space Design}
In this section we describe our attempts to build an AGD system that created explorable 3D spaces -- proto-walking simulators, in a sense. This system was never finished or named, and so we refer to it here as \textit{S1}. This was the first experiment we undertook in this area, intending to extend our existing work generating 3D games into the domain of walking simulators. 

\subsection{Prior Work}
Previously we had developed a version of a system called \angelina which designed explorable 3D spaces. However, unlike walking simulators, the emphasis of the system was on producing a more mechanically focused experience where the player navigates a maze, avoids deadly obstacles, collects objective markers, and eventually finds a goal. In addition to designing the rulesets and maze layouts, \angelina was responsible for selecting and curating 3D models, textures, music and sound effects. However, these artistic factors were not deeply evaluated by the system, compared to the rules and levels.

\angelina begins by receiving a text input describing a theme. If the input was longer than a single word, the system tries to extract a single concept using natural language processing to identify the rarity and importance of words in the phrase. This final thematic concept, like \textit{winter} or \textit{thief}, is then used to query databases of 3D models, textures, sound effects and music to design the game space. Sometimes this would be supported by other analysis - music selection, for example, was supported with sentiment analysis of the input text to try and pick an appropriate mood.

\subsection{Building S1}
Our aim with \textit{S1} was to build a system with a similar overall workflow to \angelina, but a different treatment of the various modules and processes within. There would be no obstacles or rules, as we envisioned a game without objectives or goals, and thus no need to design a maze or playtest navigation. Instead, the emphasis would be on the other parts of the design process, the ones that were less fully explored previously: choosing, arranging and placing visual and aural content in the space to create an atmosphere or mood. 

One particular aim we had was to change the way thematic content was obtained. For example, in \angelina the search topic is expanded to a list of words using a word association database, and these related words are used to search a 3D model library to find content for the game. This method's success was largely dependent on the type of query. Quite often word association sets would jump between word senses, so the word `founder' would be associated with `boat' (in the verb sense of \textit{foundering}) but also with `cult' (in the noun sense of someone who founds). This resulted in clashing sets of content that were often disorienting, but could sometimes provide interesting contrasts or puzzles for the player. \angelina gained a reputation for being `surreal' and `dreamlike', and drew comparisons to \textit{LSD: Dream Emulator}, a Japanese exploration game about strange experiences.

For S1 we wanted to avoid the surreal atmosphere and instead build a system capable of evoking specific places or moods with the spaces it designs. For example, KO-OP Mode's \textit{Weird Forest} contains trees, pathways, stone walls, and cloudy skies. These things are recognisable, placed and scaled correctly, and coherently relate to one another to evoke a particular kind of temperate forest scene. Our intention was to build a similar network of related objects, extracted from the input text, find specific models for them, and then situate them correctly in a scene. Thus instead of relying simply on word associations, and including objects arbitrarily regardless of their associations or real-world nature, we would attempt to intelligently place objects in appropriate ways.

We explored different ways of obtaining an initial word set, including using our original word associations but filtered to restrict results to noun objects. However, we recognised a need to not simply identify objects but also understand their relationship to the space. Thus, we turned to semantic databases such as ConceptNet to see if they could provide supporting information. For ideal cases, ConceptNet is passable -- for example, it can tell us that a forest might contain a camp site, dead leaves, frogs, and a stream (although it also suggests it might contain a mammoth, rain and `a creature'). Although we knew ConceptNet would not be usable in the long-term due to bias issues and incomplete data, we decided to continue prototyping to see what other issues emerged.

However, simply understanding the relationship between an object and the theme is not enough to properly design a scene. As we prototyped, we realised that important visual information was even harder to discern. For example, given a set of objects that should be arranged in a space (such as a tree, a stone wall, and cloud, for example) where should they be placed in relation to one another? How large should they be in relation to the player? How many of each object type should exist? Simply attempting to identify whether an object is a rare, one-of-a-kind object (such as a ruined castle, for example) or something that appears in the hundreds (such as a fern) is very hard to intuit, and we failed to find good, reliable data about this. Some research related to this task has since emerged, such as \cite{hodhod}, however this often relies on crowdsourcing or data gathering, which we were reticent to do as it would undermine our secondary goals of autonomy and computational creativity.

\begin{figure}
\begin{subfigure}[b]{0.45\textwidth}
\centering
\includegraphics[height=3cm]{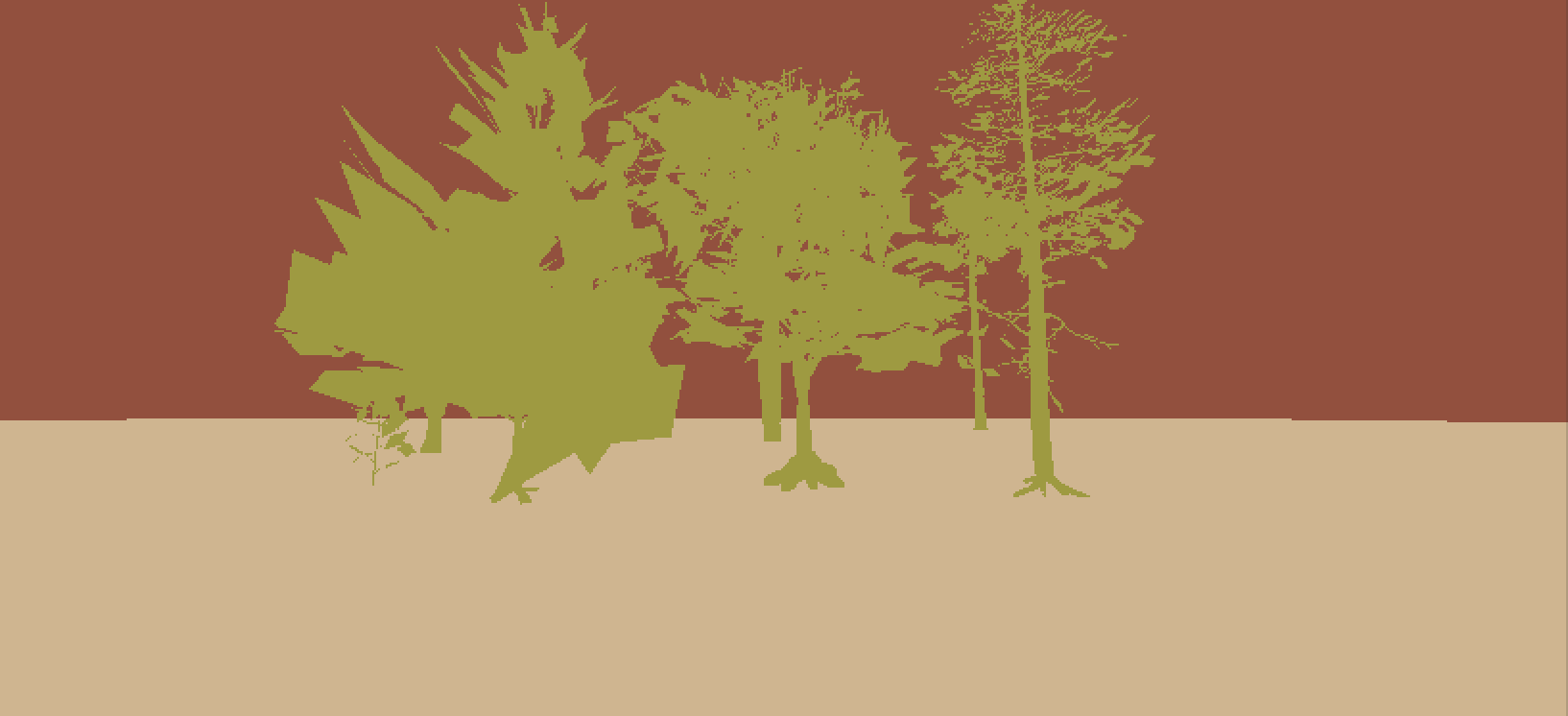}
\end{subfigure}
\begin{subfigure}[b]{0.45\textwidth}
\centering
\includegraphics[height=3cm]{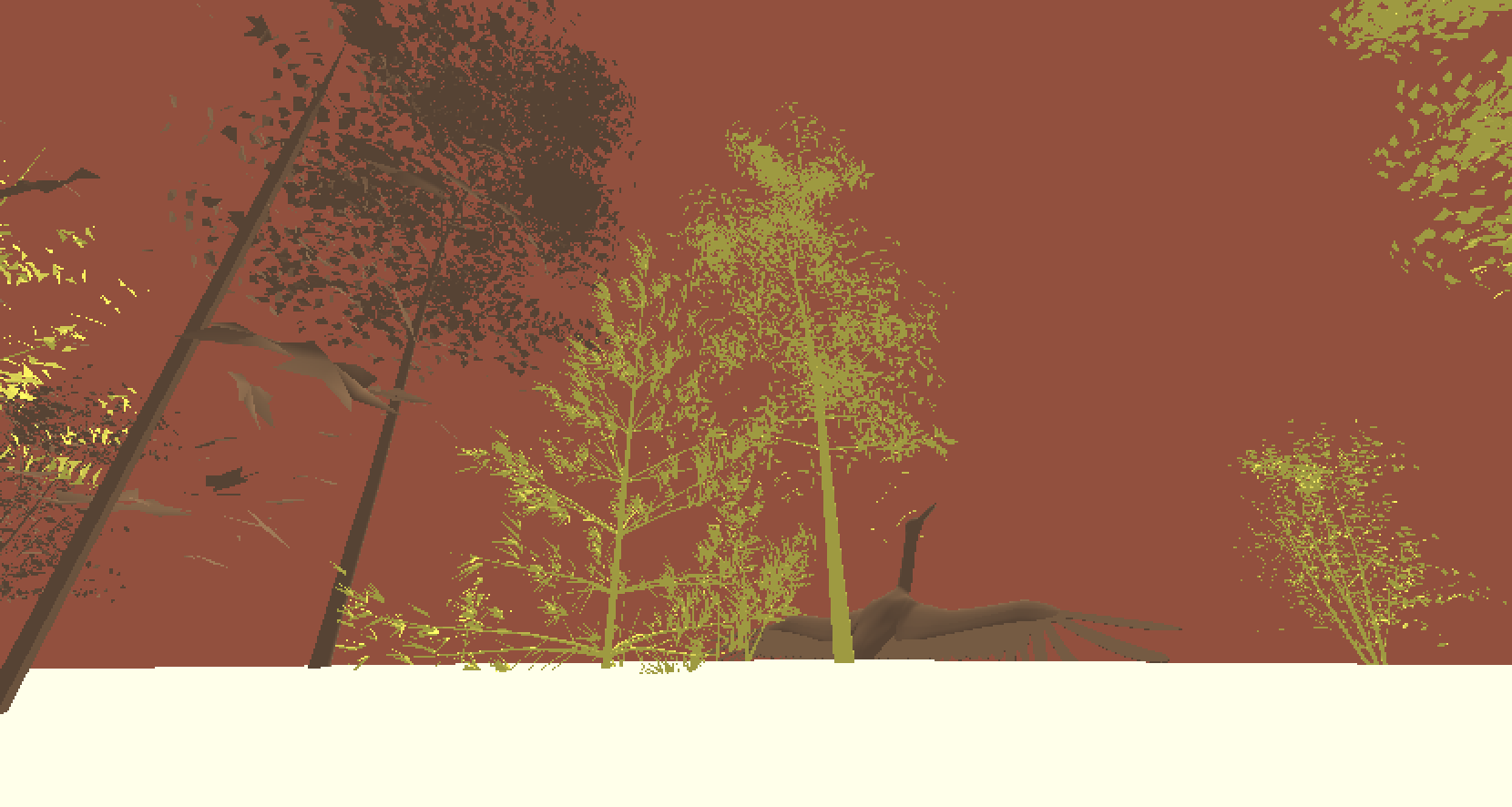}
\end{subfigure}
\caption{Two screenshots from S1. Left: a forest scene. Right: note the oversized swan model.}
\label{fig:forest}
\end{figure}

\subsection{Evaluation and Abandonment}
Despite our concerns about the process of both identifying scene components and arranging them, we did build content pipelines and test out simple scene construction. Many of the 3D models we obtained came with their own textures, which were often of different styles and quality levels, meaning they often clashed. To solve this we added a new module which found colour schemes related to the target theme, and applied flat colours to the models. Fig. \ref{fig:forest} shows two screenshots from the only surviving example of the system. The autumnal colour scheme can be seen here, as well as several different models for trees. Note also in the second image that there is a giant swan, evidence of the system's inability to decide how big a model should be relative to other objects in the scene.

We can see another issue in these screenshots which became evident at this point -- a lack of definitive way to shape the space. Not only do we not know enough about the semantic role of the 3D models downloaded, we also do not know enough about their role in the design. When designing a space like this some objects will draw the eye, others will be useful for blocking vision or shaping paths. In \textit{Weird Forest} the designer uses the trees to create dense areas, making the open spaces more inviting. The pathway stones lead the player through these open areas, and the stone walls further divide up the space. Different models play different roles in the level design vocabulary of the game. This is important information that also requires a lot of understanding about exactly what space is being designed.

Given that we were encountering these difficulties with content use even under `ideal' circumstances, using a very clear theme input (\textit{forest}), with obvious noun associations, entries in ConceptNet, and an inspiring example to compare to, we realised that this system was only going to perform worse and encounter more complicated issues once the input prompts became unconstrained. In other words, given that S1 was struggling to design a forest, how would it cope with a cathedral, the surface of Mercury, or an abstract theatrical space?

Despite this, we found the initial outputs exciting. The spaces designed by \angelina felt artificial and strange, but by removing goals, emphasising the visual content, and applying the most basic visual direction in terms of colour schemes, the spaces felt very different and more engaging. Some of our favourite outputs from the \angelina system were bugs where a model would accidentally be scaled up to a hundred times its intended size, giving an amazing sense of scale, and walking around the prototype forests felt similar. This encouraged us to keep working, and to try and build more focused systems to avoid the difficulties encountered by trying to begin with a very general approach.

\section{System 2 - Agent-Based Evaluation of Level Designs}
In other AGD systems, games are often evaluated by having an AI agent play the game. This is one reason why these systems benefit from being objective-driven and goal-directed, because it provides a focal point for game-playing algorithms to aim for. The behaviour of these agents can then be used to assess the qualities of the game -- is it too hard, too easy, does skill impact player performance, and so on. We were interested in developing agents that could use a similar approach to evaluate aspects of level designs, rather than the quality of a ruleset. 

\begin{figure}[h]
\includegraphics[width=0.9\columnwidth]{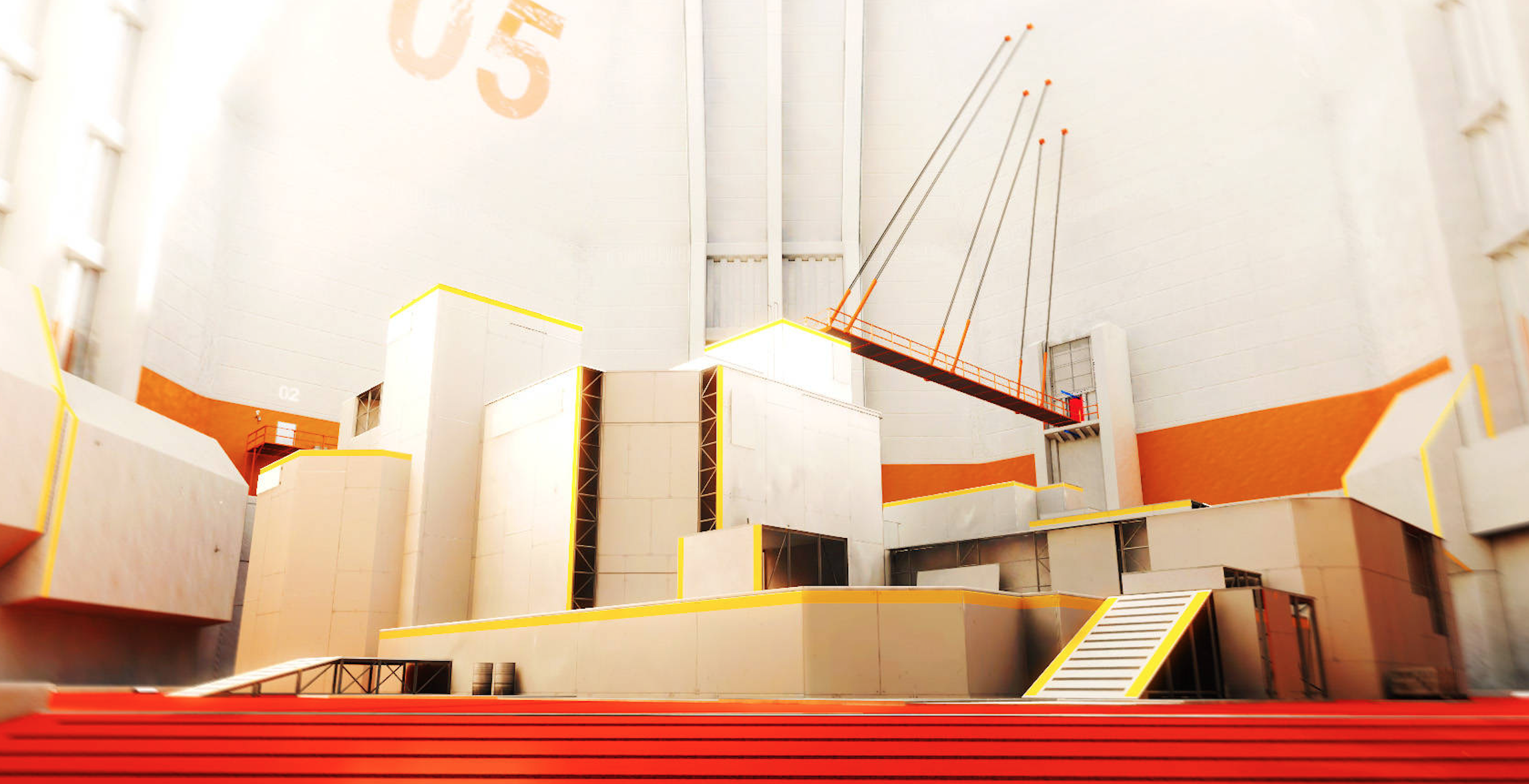}
\caption{A screenshot from \textit{Mirror's Edge}. Note the prominence of the red exit door in the distance, the sharp lines of the walkway leading towards it, and the less saturated yellows highlighting edges of platforms.}
\label{mirrorsedge}
\end{figure}

\begin{figure}[h]
\includegraphics[width=0.9\columnwidth]{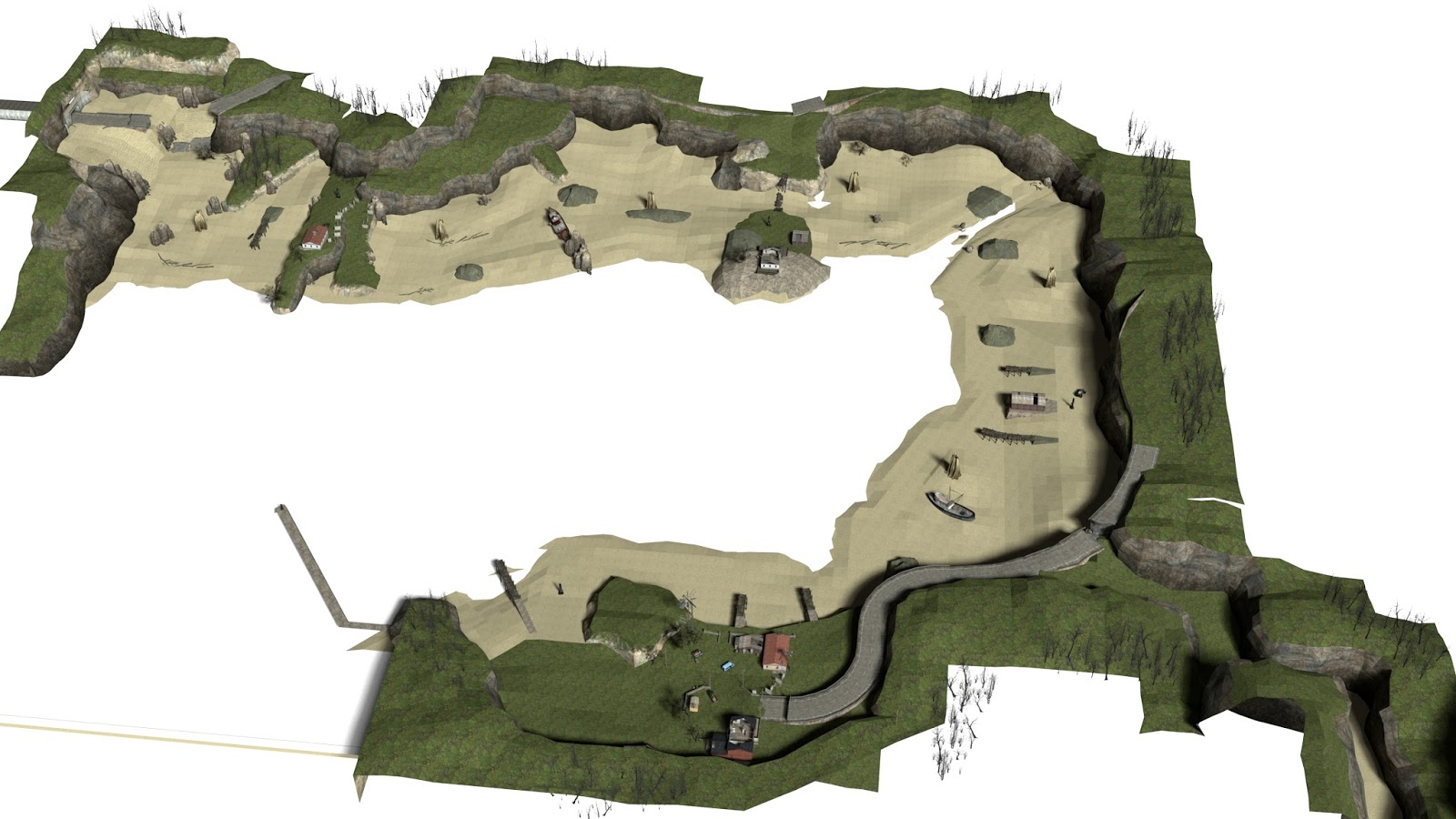}
\caption{A section of Half-Life 2, remade by Robert Yang. Open space is used to make the player's destination across the bay visible at all times.}
\label{halflife2}
\end{figure}

\begin{figure}[h]
\includegraphics[width=0.9\columnwidth]{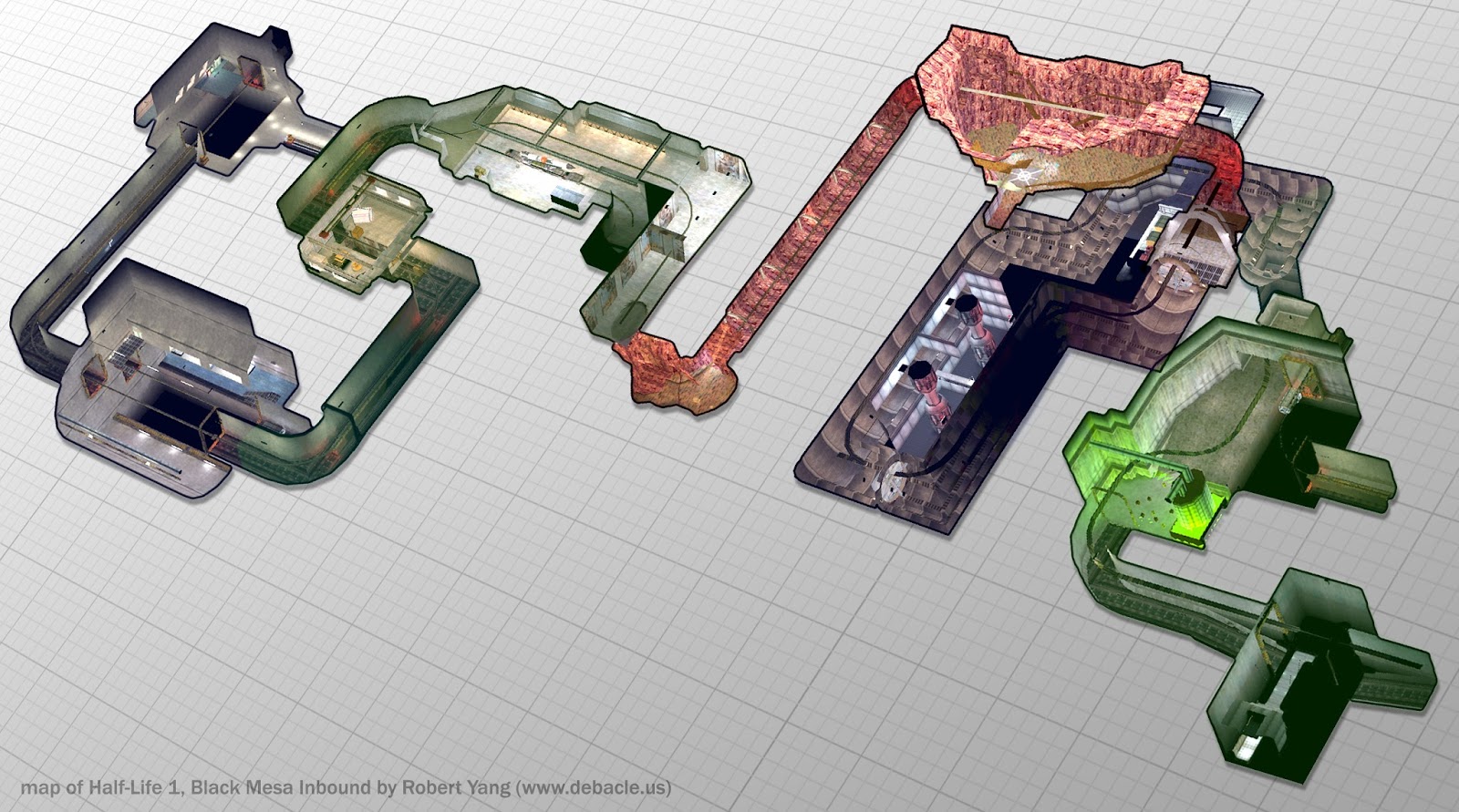}
\caption{A section of Half-Life 1, remade by Robert Yang. The snaking route of the map conceals new areas, allowing resources to be unloaded, as well as creating anticipation in the player.}
\label{halflife1}
\end{figure}

During the prototyping of S1 we read widely about level and environmental design, and studied celebrated examples such as \textit{Mirror's Edge}, \textit{Half-Life 2} and \textit{Dear Esther}, as well as writing by experts like Robert Yang \cite{yang}. Learning about the various skills involved in level design, and the different considerations that go into making immersive, directed or beautiful 3D spaces, encouraged us to study some of these ideas in more detail, and to see if we could build agents that could help evaluate some of these properties in 3D spaces. This led to a series of experiments with vision-led procedural content generation, which we will summarise here. This work was the only part of our experimentation with walking simulators that made it to partial publication\cite{exagvision}.

\subsection{Background - Level Design}
Level and environmental designers in videogames use a wide array of clever techniques to shape the player's experience and understanding of a space. This can involve subtly directing the player towards a critical path, encouraging them to explore a space, or helping the player rapidly understand the layout of a space so they can quickly react to a situation. Designers carefully manage the layout of objects, as well as colours, textures, lighting and framing, to change how the player perceives a space as they move around it.

A common recurring technique in first-person games is to frame the player's long- and short-term goals through landmarks. In \textit{Mirror's Edge}, for example, the colour red is used to indicate the way forward through a level. The player learns to look for red markers in the landscape -- such as the distant door visible in Fig. \ref{mirrorsedge}. The prominence, framing and visibility of objects is also important narratively. In \textit{Dear Esther} the game's climactic moment takes place at a radio tower, but the player is able to see this tower for the majority of the time they spend in the game, visible in the distance and identifiable by a bright red beacon at the top. 

Similarly, obstructing the view of some things can also be important. Obstructing the player's view can help set up surprises and reveals, as well as help with pacing by breaking spaces up into smaller self-contained areas. There are also technical reasons for obstruction, such as allowing sections of levels to be loaded and unloaded at runtime, or helping stop the player from noticing strange architecture that is unrealistic but necessary for the game's flow. Fig. \ref{halflife2} and Fig. \ref{halflife1} show contrasting examples -- in the first, from \textit{Half-Life 2}, the player progresses along a beach, with the open bay showing their eventual destination, giving them a sense of progress and building towards a climactic end to the chapter. By contrast, in Fig. \ref{halflife1} the player is taken through a winding path that cuts off the view of places already visited, and places yet to visit. This allows the engine to unload expensive game content when out of sight, and makes the player anticipate the next area's reveal.

\subsection{Vision-Based Level Evaluation}
We hypothesised that agent-based evaluation approaches could be adapted to take into account level design sensibilities such as the control and shaping of player vision or the framing of objects in a scene. To investigate this, we developed an evolutionary level designer whose objective function was designed to achieve specific framing and player vision goals. In order to evaluate a level, an agent plays through it, recording what it sees and when, and comparing this to the objective function's level constraints.

In our system we specify a level design task by describing a level template, consisting of an empty surface with several markers: one start marker and one end marker, which the agent uses to play the level, and one or more objective markers which represent objects we wish the player to either see or not see as they walk through the level. 

The aim of the evolutionary system is to evolve a level design, defined as a set of three-dimensional blocks with their associated positions and sizes. The system has several parameters describing the upper and lower quantities of blocks allowed in a solution, as well as maximum and minimum dimensions. 

In order to evaluate a given level design, we instantiate all of the blocks described by the design, and then place a vision-based agent evaluator at the start point. Using A* pathfinding, the evaluator walks the shortest path through the level to the end point. We attach a camera to the agent and at each frame we check each objective marker to see if it is in the frustrum of the camera. This alone is not enough to guarantee visibility, as it may be obscured by scenery, so we then cast several rays from the agent to each vertex of the objective marker model. If any of these raycasts are successful, we consider the marker visible. The fitness of a given level design is proportional to the number of markets whose visibility constraints are met (either visible or not visible) across the course of the level.

We will avoid further technical descriptions of this system here in order to focus on discussion, but for more experimentation we refer the reader to \cite{exagvision}.

\subsection{Extension and Evaluation}
\paragraph{Evolution with Complex Models} We were encouraged by early results from the system, as it was able to evolve levels that met our constraints with ease. We began extending the system to see if the same process would work with more complex objects, rather than simple blocks. Our reasoning was that because blocks have simple silhouettes, and because the system could freely scale and reshape them, it made the design task considerably easier. We wanted to see if the system would be able to achieve the same results with complex models -- for example, rocks and trees -- which it was not able to reshape.

\begin{figure}
\includegraphics[width=0.8\textwidth]{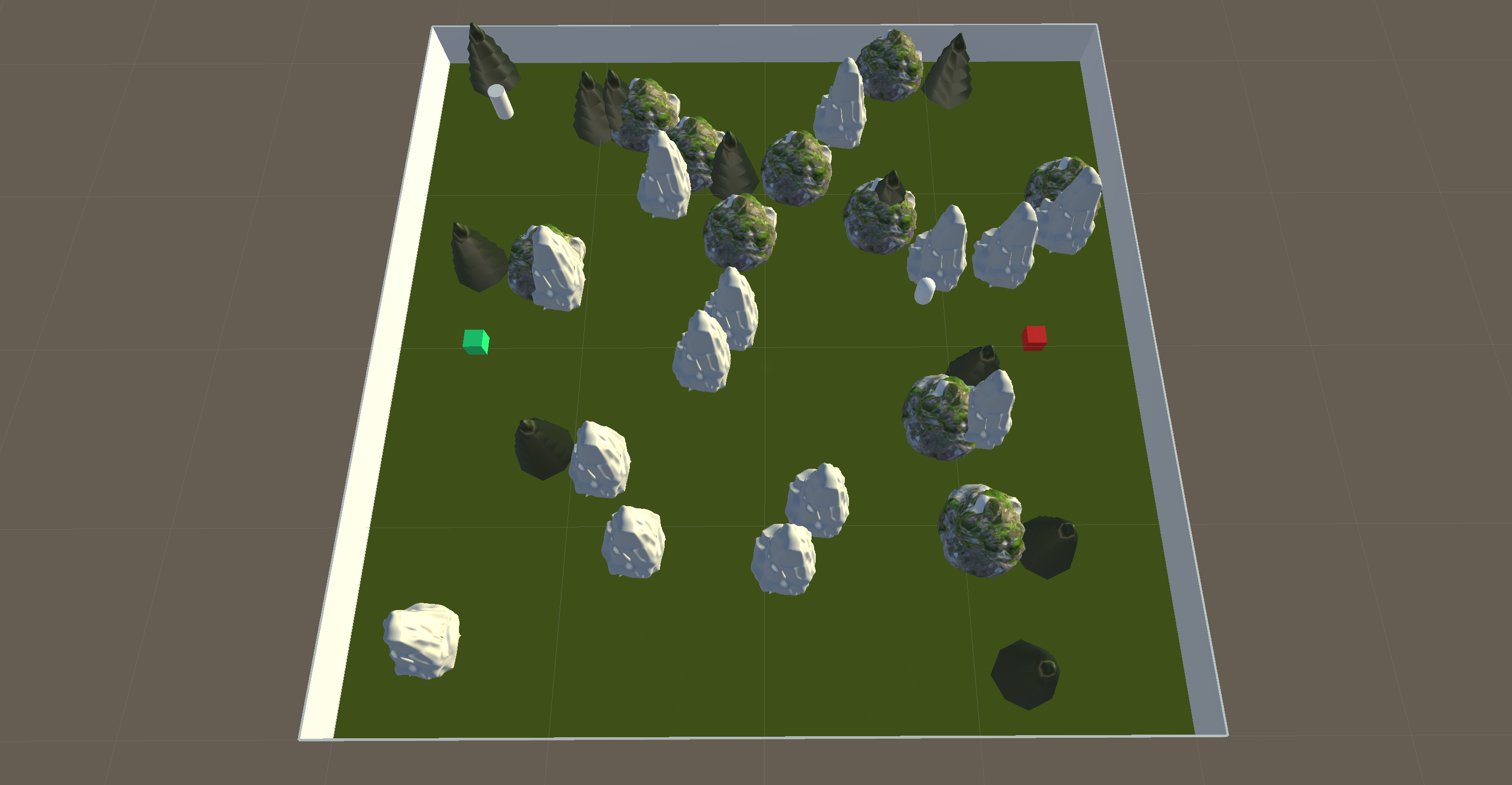}
\caption{A level evolved using tree and rock models instead of standard blocks.}
\label{fig:models}
\end{figure}

This task proved a lot harder to reliably solve with our evolutionary approach. Often very small fractions of the objective marker would be visible around the edge of a rock, and objects like trees seemed difficult for the system to use effectively to block vision. Additionally, we met again with issues related to context. Although the system could sometimes evolve good solutions, its use of models was haphazard at best - rocks and trees were randomly clustered together or scattered about the space, not behaving like rock formations or forests do in real life. Fig. \ref{fig:models} shows an example of an arrangement of rocks and trees.

It is possible that with additional constraints on the placement of objects, such as target quantities of a particular model, or metrics measuring how close particular models are to one another, that more aesthetically pleasing layouts could be evolved. However, we did not pursue this, since at the time we aimed to design systems which would be more general in their approach to space design, and thus we would not be able to derive these design metrics automatically or know which models the system would be using. In retrospect now, we believe this would be worth pursuing, with the aim of building a system that uses a pre-selected set of models. The evolutionary system could be tuned to suit that model set, and then potentially become a useful design tool. As we will see in the next section, we ultimately move to a more specialised, less general approach in the end anyway.

\begin{figure}
\begin{subfigure}[b]{0.45\textwidth}
\centering
\includegraphics[width=\textwidth]{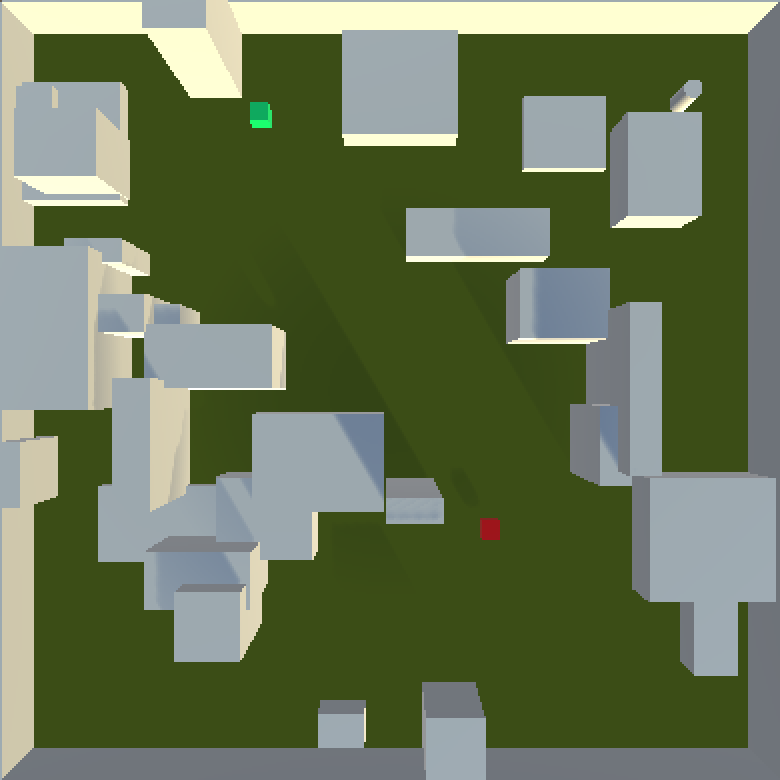}
\caption{Design constraints: both markers visible.}
\end{subfigure}
\begin{subfigure}[b]{0.45\textwidth}
\centering
\includegraphics[width=\textwidth]{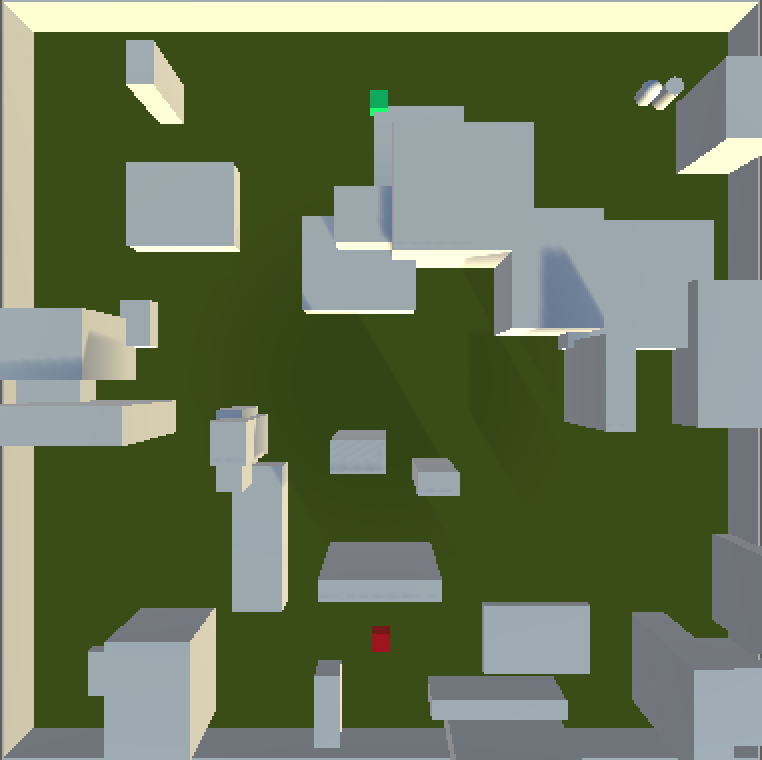}
\caption{Design constraints: top marker visible only.}
\end{subfigure}
\caption{Two results from the evolutionary level designer. The objective markers are the smaller coloured cubes. Agents start in the bottom-left and walk to the top-right.}
\label{visionresults}
\end{figure}

\paragraph{Curious Exploration} In the evolutionary system described above, the agent has total knowledge of the space it is pathing through, and walks directly to its goal. This is an unrealistic model for several reasons. Most obviously, players do not know about spaces in advance, and exploring them from a first-person perspective limits their ability to path and plan. Furthermore, players often do not have explicit goals either. This is why the language of level design is so effective - it leads the player to make natural-feeling discoveries without telling them what to do.

Building on our work evolving levels, we began prototyping a new type of agent that would not know the layout of a map in advance, and instead explore based on what it could see in its surroundings. A grid of points is overlaid on top of the map, and the agent builds up an understanding of how these points connect to one another over time as it explores and sees more of the map.

\begin{figure}\includegraphics[width=0.8\textwidth]{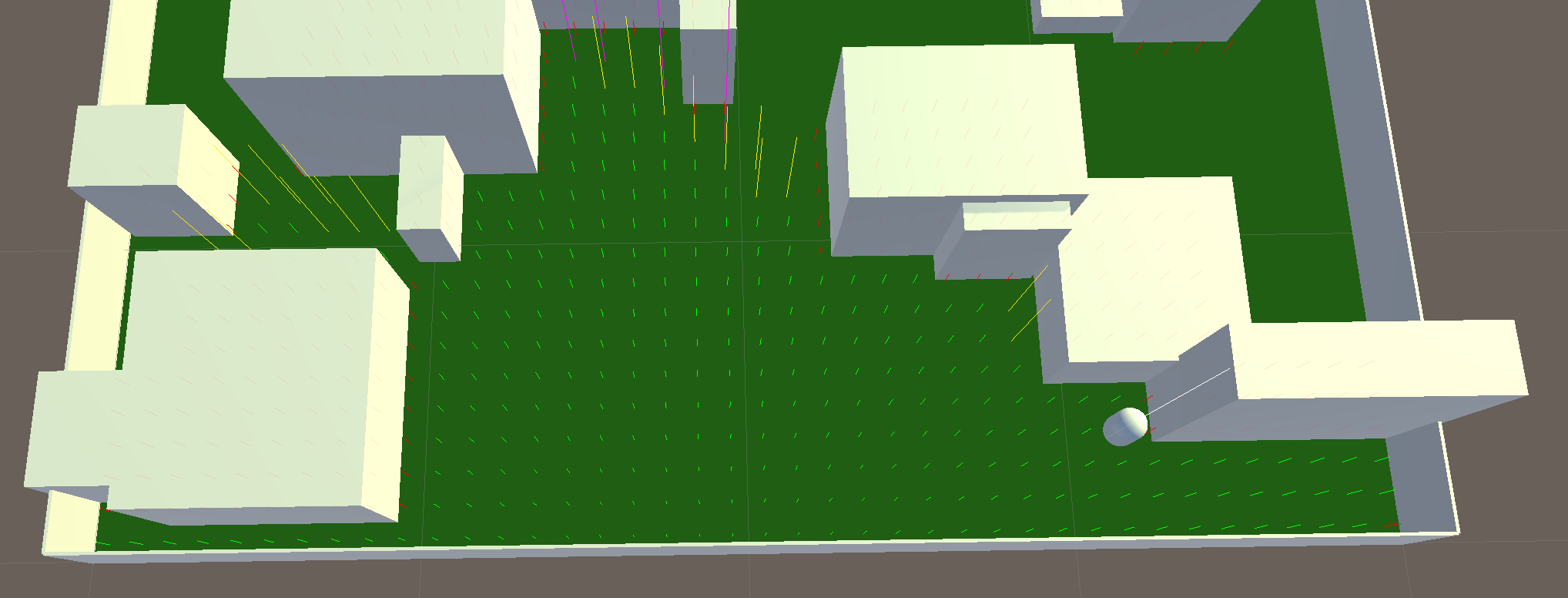}
\caption{The debug view of the curious exploratory agent.}
\label{fig:vision}
\end{figure}

Fig. \ref{fig:vision} shows a screenshot of the agent assessing its environment. Markers in the ground show the agent's awareness - green markers indicate a point in the world that is currently visible. Yellow markers represent points that the agent believes to be empty, and are adjacent to currently-visible points. These are points the agent is curious about and will move towards. Magenta markers represent points which were yellow in the past, but are currently neither visible nor adjacent to visible points. 

We struggled to get this agent to reliably explore levels, particularly low-fitness levels. The agent would often get confused about whether it could reach a point in the world (because it had no ground truth for pathfinding, only estimations based on vision) and we failed to find a way to fix this. We were still working on this agent at the time we stopped work on the project as a whole.

\section{System 3 - Constrained 3D Landscape Design}
One lesson we took away from our vision-based experimentation was that many of the tasks we were interested in studying in the discipline of level design were tightly related to one another, and that they were also very easily disrupted by changing other aspects of the game design. For example, making a change in the kind of 3D model we used to evolve level designs could completely change the sensitivity of the vision-based agent, making it much harder to design levels with visibility constraints.

We see similar coupling in other AGD systems, for example between rulesets and level designs, where a small change in a ruleset invalidates level designs that depended on particular rules to enable their solutions. If this interdependency effect is too strong, one solution is to fix part of the solution space so that the AGD system can focus on solving the remainder of the problem. This sacrifices novelty and possibility space size in exchange for tractability and stability.

We decided to take a similar approach with System 3, by constraining the system to a specific design problem, fixing certain aspects of the game's design in advance. Our decision was also influenced by discussions at the Computational Modeling in Games workshop held at Banff in 2016. One workgroup, led by Kate Compton, discussed the design of real-world spaces such as theme parks and public gardens, and their relationship to human memory. We found this inspirational, and the idea of making small garden-like spaces to explore was very appealing.

\subsection{Early Prototyping}
We experimented by building a few early prototypes. One used Perlin noise discretised into a few different height levels to create spaces with varying elevation, but this approach alone felt too undirected. We then tried partitioning a large plane into a Voronoi map, and randomly walking across the map to place solid ground tiles. Map tiles that were not walked on were turned into water, dividing up the space. This gave enough texture to the space to decorate it -- fencing was placed around some of the borders between water and land, with appropriate plant life (such as trees and lilypads) placed on tiles.

\begin{figure}
\begin{subfigure}[b]{0.45\textwidth}
\centering
\includegraphics[width=\textwidth]{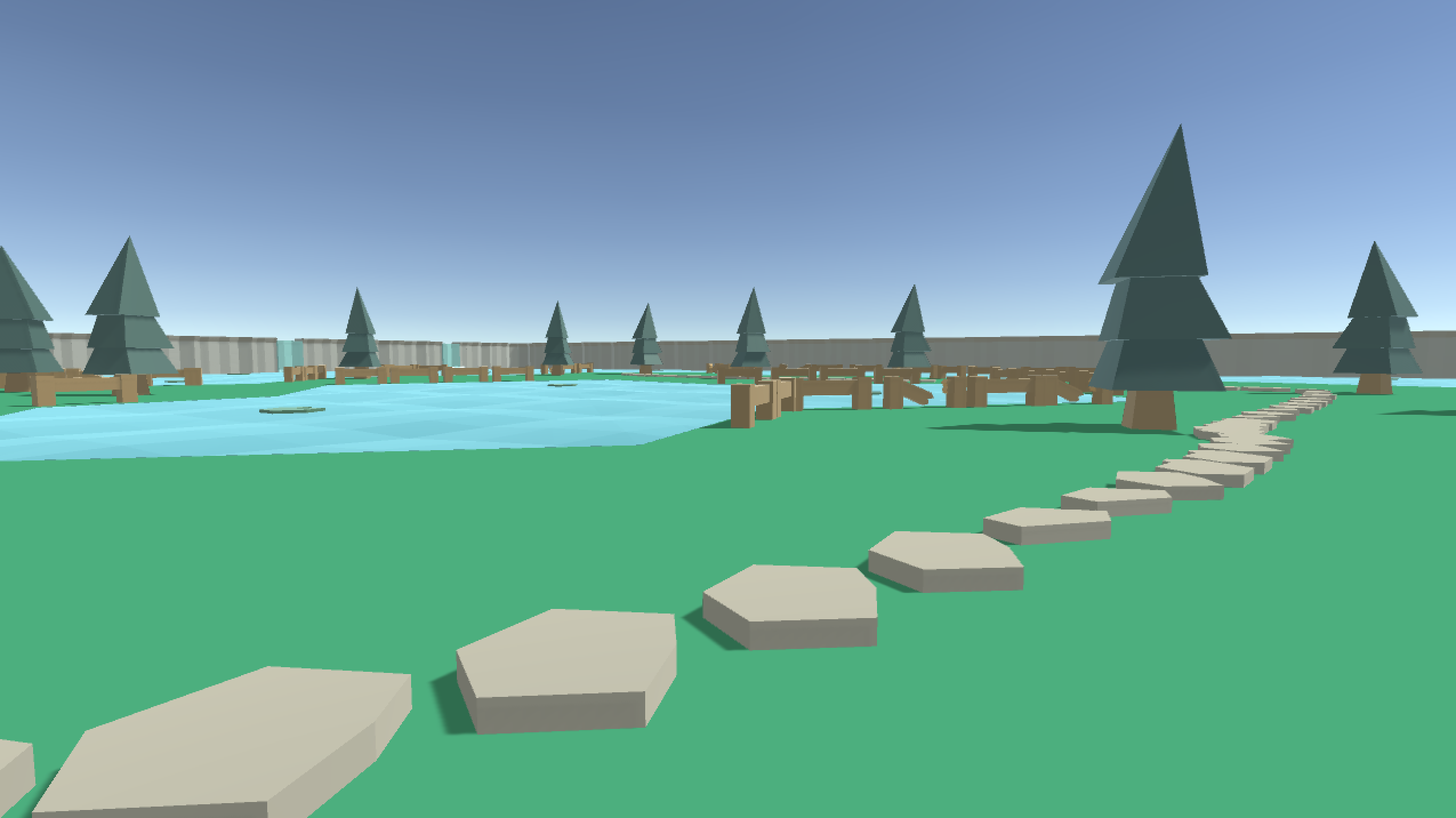}
\caption{Player perspective.}
\end{subfigure}
\begin{subfigure}[b]{0.45\textwidth}
\centering
\includegraphics[width=\textwidth]{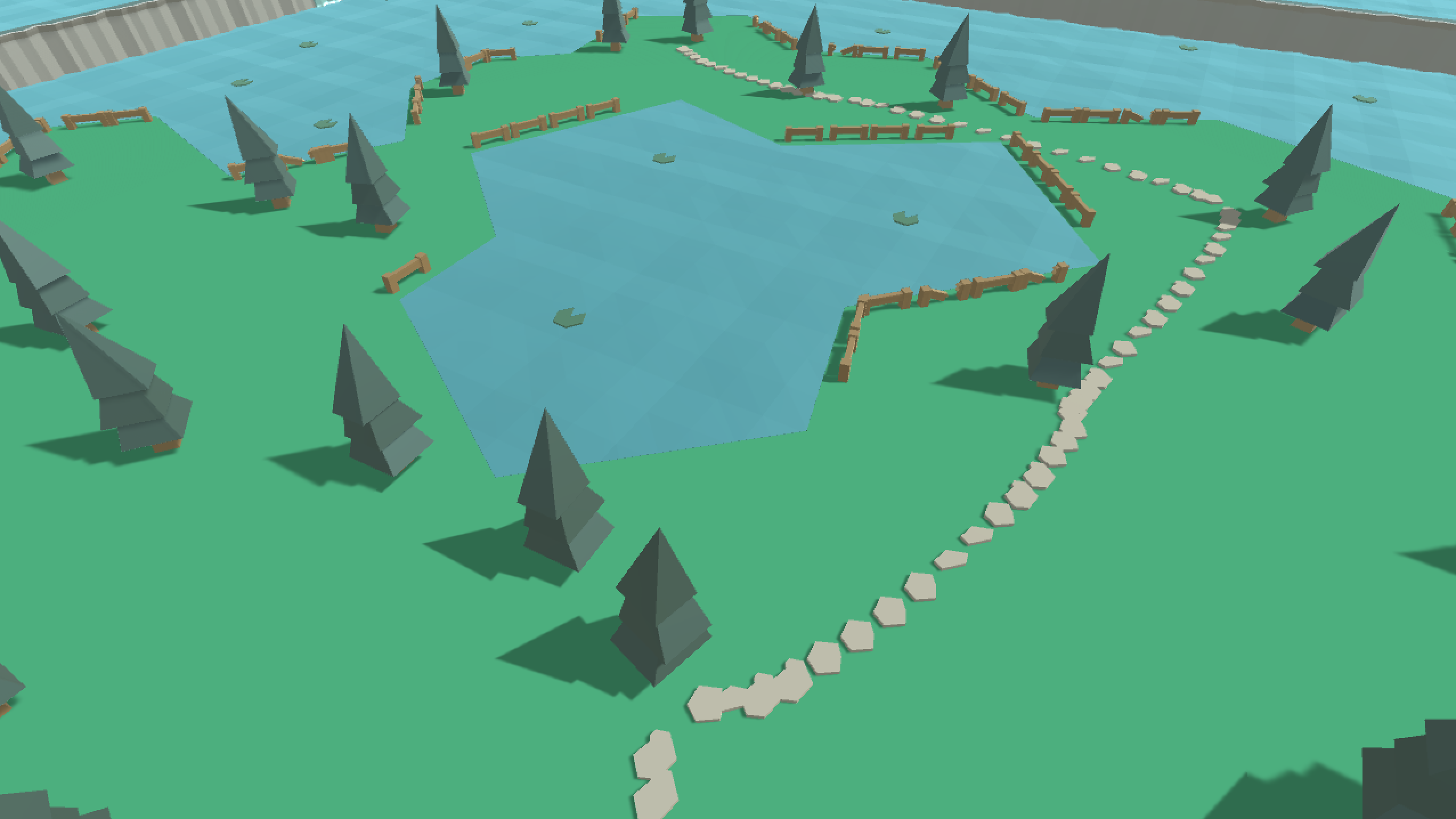}
\caption{Overhead shot.}
\end{subfigure}
\caption{Screenshots from an early prototype of System 3.}
\label{fig:mudeford1}
\end{figure}

These spaces were more visually pleasing, and some iterations would produce interesting rivers, bays or inlets that were particularly nice to walk around. The spaces felt directionless, though, so we added a path to guide the player across the space. The path was laid by picking tiles at opposite ends of the landmass and then running a pathfinding algorithm across the space, placing stone path models along the route. Fig. \ref{fig:mudeford1} shows a screenshot from this early prototype.

This produced pleasant spaces, but the paths felt \textit{too} directed -- they guided the player fairly abruptly to nowhere in particular, and their route was not based on anything we expected the player to see -- it didn't attempt to go by a particular route, see a particular landmark, or achieve any goal. The prototype was useful in helping us feel out the design of spaces, but we were unsure how to create interesting landmarks without adding them in by hand, thus reducing the variation of the system's output.

\subsection{Mudeford}
We developed the prototype further for \textit{PROCJAM 2016}, a game jam focused on procedural generation. Removing the path generation, we added in different types of model to include some larger pieces such as rock formations, and made the overall level size larger. At one end of the space we placed the player, and at the other end a campsite, acting as a little surprise for the player and a potential ending point for play. 

The initial results were pleasant to explore, but there was little to differentiate each generated island. Islands weren't bad or good, there was little in the way of a focal point, and not much opportunity for an AGD system to create something interesting or different. The best islands offered occasional lakes or rivers that were nice to walk past, but it felt hard for us to express exactly why this was nice in any reusable metric - it didn't mesh with any of our existing experiments into evaluation, for example.

While working on the prototype, someone compared the game to their daily dog walks. This inspired us to build this into the game -- we added a dog  NPC that follows the player around. Originally the dog would wander in the approximate direction the player was moving, but it was very easy to lose track of the dog's location, causing the player to look around, confusing the dog's pathfinding, causing a feedback loop. We rewrote the dog's behaviour so it always walks back into the camera frustrum, which reduces player panic and simplifies the route back into visibility.

The resulting prototype, \textit{Mudeford}, can be played online\footnote{cutgarnetgames.itch.io/mudeford}. The response to the prototype was generally favourable as a simple game jam entry, although it is a very early prototype in terms of being an AGD system. 

\subsection{Abandonment}
The Mudeford prototype was well-received, even receiving coverage in \textit{Kill Screen} as a relaxing few minutes of distraction. However, after the jam we stopped work on both System 3 and System 2 experimentation. Although the initial results were pleasing, it felt hard to know where to go next -- the prototype did not offer a lot of obvious opportunities to hand responsibility over to the AGD system in ways other than varying the placement of a few object types, and we didn't possess the skills required to do more complex spatial design such as landscaping or lighting. Without a firm understanding of these skills ourselves, we felt unable to design a system that could undertake this work.

At the same time, we were developing new ideas about the role of AGD systems in the design process with regards to computational creativity \cite{continuous}. We had originally intended to explore these through systems which designed walking simulators and 3D spaces. However, given that our goals for computational creativity research were already complicated and experimental, we felt a need to shift to a more constrained, understood design problem to provide a stable foundation to build on. We discuss this further in the following discussion section.

\begin{figure}[h]
\includegraphics[width=0.8\columnwidth]{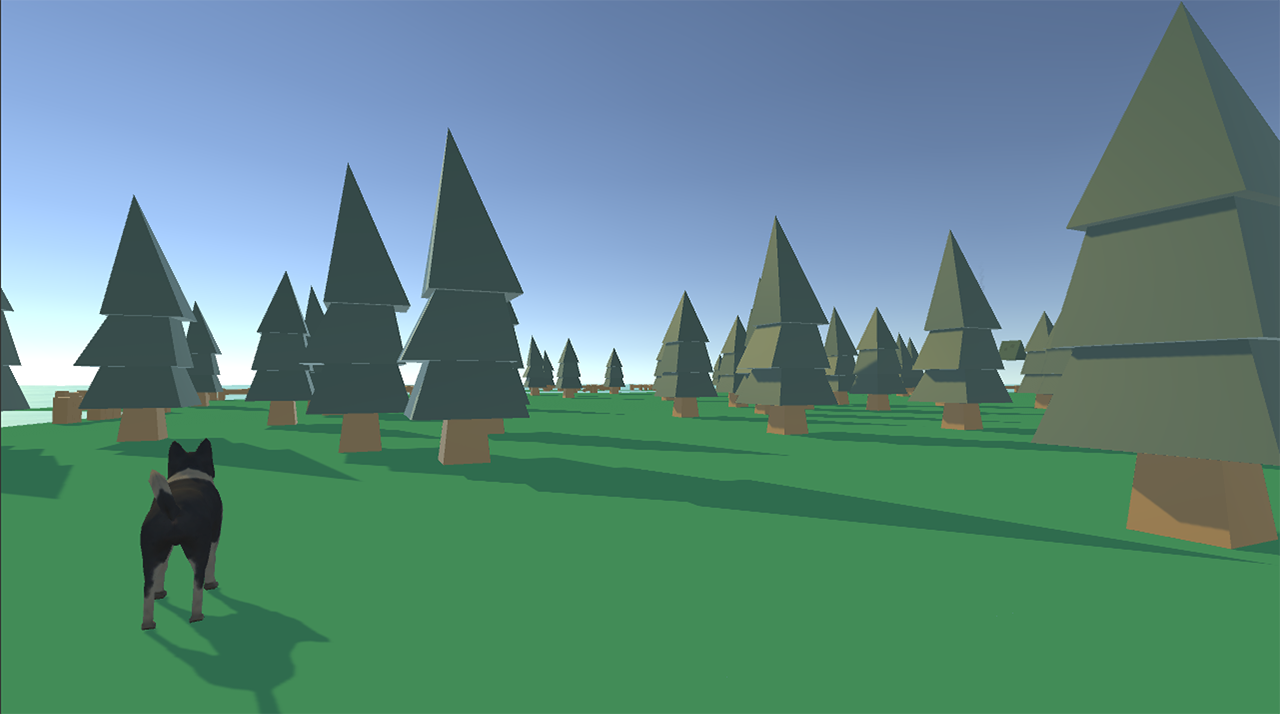}
\caption{A screenshot from the \textit{Mudeford} prototype. The dog character is visible on the left of the shot.}
\end{figure}

\section{Discussion}
\subsection{Lessons}
\paragraph{Generality Requires Experience}
Aiming for generality is often useful - it challenges us to think deeply about the problem domain, and to find unusual solutions. It helps us divide problems up and see what separates one task from another. However, in building these systems we found generality a source of many problems. The reason for this, we feel, is that thinking about a problem in the general case requires a good understanding of how to solve specific instances of the problem. For the design of AI which created game rules and systems, for instance, we felt we had some experience with designing games ourselves, which helped us. For environmental design, even with our attempts to learn and practice, we were nowhere close.

Constraints helped - building more specific systems, such as \textit{S2} to test particular ideas, or \textit{S3} to test particular design domains. We recommend constraining these problems down further and further until they become tractable enough. We burned out too fast on the more general approach in \textit{S1} which left us with less energy for the more specific follow-up projects.

\paragraph{Vision Is Underused}
Despite all of the problems we encountered in the project, we were surprised at how effective vision was in some of our evolutionary experiments. The success of our marker visibility experiments was encouraging, and we believe more complex AI agents could be built with a variety of vision-based skills for evaluating spaces, especially given the various advances in machine vision over the last decade, which can be even more effective in a game context since they have access to the `ground truth' of the world in terms of the real placement of objects.

We still believe that strong agents, that can measure and respond to the world as they perceive it, will be very useful in building sophisticated walking simulator AGD systems. Unfortunately it is difficult to get access to example human-designed 3D spaces to explore. Vision-based agents could be trained purely on camera input, but that feels like a wasted opportunity given how much useful data could be extracted from the software itself. We would recommend finding funding to pay a range of designers of various experience levels to make micro-spaces\footnote{What could it cost, Michael? Ten dollars?}, in the style of KO-OP MODE's \textit{Weird Forest} tutorial, to provide a set of example spaces for researchers working in the area.

\paragraph{Framing Devices}
After adding the dog to \textit{Mudeford}, we realised that it could potentially be a very powerful tool for an AI designer. The dog can be used to draw player attention in a more intentional way - we hypothesised that the player would follow the dog's attention, allowing the dog to point out landmarks, lead the player to areas they may have not noticed, all without explicitly telling the player what to do or where to go. This is somewhat similar to the function of AI companions such as Elizabeth in \textit{Bioshock Infinite}. 

We believe that building in devices such as this, or explicitly marked paths, can help the AGD system frame its work in stronger ways, without having to rely on the extremely subtle clues that many level designers rely on. This might help bootstrap the AI's lack of human perception, and give new ways for an AGD system to communicate with the player.

\subsection{Reflections}
It was both ironic and sad to abandon this line of work, originally started to buck a trend in automated game design, only to return to more traditional AGD domains. However, we feel it has paid off in the long-run -- we are happy with the work we've been able to do since and feel it has also helped broaden and progress ideas about AGD, albeit along different lines. We still feel that the study of 3D spaces, environmental design, and walking simulator design, are all very important ideas for AGD to address, however. We were not the right people to do this work at the time, but we learned a lot from our experimentation, and that helped us grow our perspective on game design as a whole, as well as help us form ideas about future research. We would still like to return to these problems some day.

It must also be said that we might not have abandoned these projects so completely were it not for the pressures of the academic career path. This work came at a time when the author was writing up a PhD, completing a Postdoc, navigating job searches and applying for funding. While the rate or quality of publications was not a concern, rationing time certainly became one. Working in this space required a lot of studying new techniques and skills, many of which (such as lighting or environmental design) were hard to find good resources about. Any university employee or student reading this has probably experienced similar, but it feels worth pointing out, in a paper reflecting on past work, that the conveyor belts, deadlines and systems we are trapped in frequently squash work and new lines of research simply through the pressure they put on us.

Separately from the acknowledgements section of this paper, we would also like to underline here the enormous impact of game developers -- a term which we use to describe anyone working in games, rather than programmers or designers -- as well as game critics and journalists on this work. The impact of people \textit{outside} academia on this work, as is true for most of our work in AGD research, was more considerable than the impact of people \textit{within} academia\footnote{An earlier version of this paper erroneously identified Robert Yang in this group - Robert is a professor at NYU and was during this work too. This is an oversight on my part, although it is interesting to think about how a lot of valuable communication between academics happens outside of conferences -- through blogs, social media, and more -- which was also one of the motivations for this paper. Thanks to Matthew Guzdial for pointing this mistake out}. We say this not as a slight on our colleagues but rather to emphasise the enormous, often uncited and unrecognised, work done by people to help games research. This work would literally not be possible without the efforts of people who wrote about and made games in the `walking simulator' genre; people who go to great lengths to educate people about intimidating and challenging disciplines within game development; people building inclusive and accessible spaces for people to share challenging and experimental work. Broadening a research field is more than just broadening the questions it asks -- it also means broadening the people participating in it, which is something AGD must also strive to do in the future.

\section{Conclusions}
In this paper we described several projects, undertaken between 2014 and 2016, that attempted to build an AGD system which generated walking simulators. We looked at the origins of the work, inspired by a paper discussing the problems and limitations of current AGD research, and saw how our attempts to move from an old AGD system into this space stalled because of large design challenges. We then looked at more specific experiments to design vision-driven agents and constrained space designers were more successful, but ultimately stalled and were abandoned due to shifting research priorities.

We are proud of this work and enjoyed making all of the prototypes and experiments listed in this paper. In writing this, we discovered many things we had forgotten working on, and being able to write a connecting story that contextualises both the systems and our thought processes at the time was extremely valuable. We hope that, even though many of these ideas are dated and the systems they describe defunct, the recording of the work and its position in the history of AGD research proves useful to those working on similar problems in the future. We believe now, as we did in 2014, that broadening this field is the key to making it impactful, helpful and exciting.

\section{Acknowledgements}
I owe a debt of gratitude to so many people for helping me throughout the period in which this work was done. Special thanks to Simon Colton, without whose support this experimental and weird work would never have been attempted in the first place. Thanks to Gillian Smith for always wanting to do something new, and working with me to write the paper that kicked this off. Thanks to Jamie Woodcock and Mark Johnson for their infinite patience and love. Thanks to Tom Betts, Tom Francis, Alan Hazelden, Guilherme T\"ows, as well as Feral Vector, Videobrains, and all the wonderful people in the UK games scene who were so welcoming and eager to teach and show me things. Thanks to all the people in the academic community who have shown me kindness and enthusiasm. Thanks to Chris Donlan for everything. And of course, thanks to the Royal Academy of Engineering for supporting this reflective writeup and my ongoing research, despite my track record being made up of weird stuff like this. If I didn't remember to thank you here, I'll buy you a cake sometime to make up for it.

\bibliographystyle{ACM-Reference-Format}
\bibliography{agd}

\end{document}